\documentclass[sigconf]{acmart}
\AtBeginDocument{%
  }

\copyrightyear{2025} 
\acmYear{2025} 
\setcopyright{cc}
\setcctype{by}
\acmConference[HAIPS '25]{Proceedings of the 2025 Workshop on Human-Centered AI Privacy and Security}{October 13--17, 2025}{Taipei, Taiwan}
\acmBooktitle{Proceedings of the 2025 Workshop on Human-Centered AI Privacy and Security (HAIPS '25), October 13--17, 2025, Taipei, Taiwan}\acmDOI{10.1145/3733816.3760760}
\acmISBN{979-8-4007-1905-9/2025/10}

\usepackage{dirtytalk}
\usepackage{graphicx}
\usepackage{subcaption}

\usepackage{etoolbox}

\begin{document}

\title[Exploring the Alignment of Human and LLM Perceptions of Privacy]{LLM-as-a-Judge for Privacy Evaluation? Exploring the Alignment of Human and LLM Perceptions of Privacy in Textual Data}

\author{Stephen Meisenbacher}
\email{stephen.meisenbacher@tum.de}
\affiliation{%
  \institution{Technical University of Munich \\ School of Computation, Information and Technology}
  \city{Garching}
  \country{Germany}
}

\author{Alexandra Klymenko}
\email{alexandra.klymenko@tum.de}
\affiliation{%
  \institution{Technical University of Munich \\ School of Computation, Information and Technology}
  \city{Garching}
  \country{Germany}
}

\author{Florian Matthes}
\email{matthes@tum.de}
\affiliation{%
  \institution{Technical University of Munich \\ School of Computation, Information and Technology}
  \city{Garching}
  \country{Germany}
}

\renewcommand{\shortauthors}{Meisenbacher, Klymenko \& Matthes}

\begin{abstract}
Despite advances in the field of privacy-preserving Natural Language Processing (NLP), a significant challenge remains the accurate \textit{evaluation of privacy}. As a potential solution, using LLMs as a privacy evaluator presents a promising approach -- a strategy inspired by its success in other subfields of NLP. In particular, the so-called \textit{LLM-as-a-Judge} paradigm has achieved impressive results on a variety of natural language evaluation tasks, demonstrating high agreement rates with human annotators. Recognizing that \textit{privacy} is both subjective and difficult to define, we investigate whether LLM-as-a-Judge can also be leveraged to evaluate the privacy sensitivity of textual data. Furthermore, we measure how closely LLM evaluations align with human perceptions of privacy in text. Resulting from a study involving 10 datasets, 13 LLMs, and 677 human survey participants, we confirm that privacy is indeed a difficult concept to measure empirically, exhibited by generally low inter-human agreement rates. Nevertheless, we find that LLMs can accurately model a global human privacy perspective, and through an analysis of human and LLM reasoning patterns, we discuss the merits and limitations of LLM-as-a-Judge for privacy evaluation in textual data. Our findings pave the way for exploring the feasibility of LLMs as privacy evaluators, addressing a core challenge in solving pressing privacy issues with innovative technical solutions.
\end{abstract}



\maketitle

\section{Introduction}
The study of privacy-preserving Natural Language Processing (NLP) has grown in significance as the rise of Large Language Models (LLMs) and their myriad applications likewise continue to advance \cite{sousa2023keep}. With this rapid progression of powerful language technologies, the trend of increased data collection and usage to power the underlying NLP models appears to be at odds with growing concerns over privacy and its protection \cite{klymenko-etal-2022-differential}. Researchers have responded to the call for privacy protection in full force, developing a diverse array of privacy-enhancing solutions in the collection and processing of natural language data for language model training \cite{sousa2023keep, 9592788}. Among these are popular examples such as text anonymization \cite{9298747}, Differential Privacy based methods \cite{hu-etal-2024-differentially}, and encryption \cite{yan2024protectingdataprivacylarge}.

A persistent challenge in the development of privacy-preserving NLP solutions is the \textit{evaluation of privacy} \cite{li-etal-2024-privlm}. As the definition of \textit{privacy} is both subjective and varied \cite{10.1145/3531146.3534642}, quantifying privacy (protection) often takes the form of some privacy \textit{proxy}, such as plausible deniability, simulated attack success rate, or semantic distance approximations \cite{SHAHRIAR2025104358}. While such proxies are useful for capturing aspects of the elusive privacy notion, they are merely approximations and do not necessarily reflect human opinions on what privacy preservation in textual data entails. This represents a significant gap, as the ultimate purpose of any privacy-preserving NLP solution is the safeguarding of individual privacy. 

Concurrently in the NLP field, the usage of LLMs as evaluation models has starkly increased \cite{li2024llmsasjudgescomprehensivesurveyllmbased}, particularly following early demonstrations of their ability to match human perceptions of complex topics such as text coherence or answer correctness \cite{10.5555/3666122.3668142}. This general approach of evaluation, termed \textit{LLM-as-a-Judge}, reduces the burden of conducting large-scale human evaluations, where human judgments are prone to errors, biases, misunderstandings, and time and budget limitations \cite{bojic2023hierarchical, elangovan-etal-2024-considers}. On the other hand, many works have shown that LLMs, such as from OpenAI, Google, Meta, or Anthropic, are very strong in providing reliable and scalable evaluations for any number of criteria, for example, in annotating data or rating the generated outputs of other LLMs or methods \cite{bavaresco2025llmsinsteadhumanjudges}, even in subjective tasks \cite{pan-etal-2024-human}. In some cases, the LLM-provided ratings can achieve over 80\% agreement with human raters \cite{10.5555/3666122.3668142}, showcasing the promise of LLM-as-a-Judge.

The feasibility of LLM-as-a-Judge has not been investigated in the context of privacy, namely, whether LLMs can provide an objective assessment of the privacy sensitivity of textual data. Moreover, it is currently unknown whether such an assessment aligns with human perceptions of privacy in text, which is largely exacerbated by a lack of research on empirically measuring human perceptions of text data privacy \cite{tesfay2016challenges}. We address these gaps in this work, guided by three overarching research questions:

\begin{itemize}
    \itemsep 0em
    \item[\textbf{RQ1.}] How do LLMs judge privacy in textual data, and how does such an assessment differ across model families and sizes?
    \item[\textbf{RQ2.}] How can we systematically measure the privacy perceptions of humans, specifically with textual data, and to what degree do humans agree on a notion of privacy in text?
    \item[\textbf{RQ3.}] What does the alignment of privacy perceptions in text between LLMs and humans teach us about the future of privacy evaluation in privacy-preserving NLP?
\end{itemize}

We not only explore the potential for using LLM-as-a-Judge for privacy evaluation in text data, but we also measure the alignment of LLMs with human perceptions. Our results show that LLMs are quite capable of assessing privacy sensitivity levels in texts, especially when given a structured privacy scale. General agreement between LLMs is observed within similar model families and model sizes, with larger LLMs showing the most cohesive results. When compared to the \textit{average} human rating of privacy, LLMs show much higher agreement as opposed to inter-human agreement, whereas pairwise LLM-human agreement levels are significantly lower. Our results demonstrate the promise of LLM-as-a-Judge for privacy evaluation, as well as highlight the intricacies of human perceptions of privacy, where capturing a general privacy notion can be greatly different than representing individual privacy opinions.

We make the following contributions to the intersection of privacy-preserving NLP and human-centered AI privacy:
\begin{enumerate}
    \item To the best of our knowledge, we are the first work to investigate the feasibility of LLM-as-a-Judge for privacy evaluation in textual data.
    \item In addition to testing 13 different LLMs, we conduct a survey with 677 humans spanning the globe, allowing for a joint analysis of LLM and human perceptions of text data privacy.
    \item We publicly release our LLM prompts, survey application, and resulting datasets, in order to allow for follow-up studies on LLM-as-a-Judge for privacy. This repository is found at: \url{https://github.com/sjmeis/privacy-judge}
\end{enumerate}

\section{Background and Related Work}
\paragraph{Privacy-preserving NLP}
Privacy-preserving NLP (or, PPNLP) centers around finding ways in which NLP can be conducted while respecting the privacy of the individuals whose data is being used to develop NLP models and systems \cite{habernal-etal-2023-privacy,9592788}. While the goal is clear, the quantification of successful privacy preservation can be very difficult, rooted in the fact that privacy is not well-defined or even widely agreed upon \cite{10.1145/3531146.3534642}. Rather, language data is subjective, unstructured, and highly contextual \cite{10.1145/3613905.3643983}, making the evaluation of privacy in NLP far from straightforward, particularly in the decision of \textit{what} makes a text private \cite{tesfay2016challenges,10320187}. 

Previous work has established the various angles from which PPNLP can be tackled \cite{10.5555/3666122.3668142,SHAHRIAR2025104358}. This includes data safeguarding and obfuscation, such as with Differential Privacy \cite{klymenko-etal-2022-differential} or text anonymization \cite{lison-etal-2021-anonymisation}, as well as private model training and inference \cite{yan2024protectingdataprivacylarge}. Although the technical methods can differ greatly across the field of PPNLP, these approaches are united in the goal of providing some form of privacy protection, whether theoretical, empirical, or both, to the users behind the text data being shared or used.

An essential part of any privacy-preserving solution in the context of NLP is its privacy evaluation. Lacking standard definitions of text privacy has led to a divergence of evaluation approaches in PPNLP \cite{meisenbacher-etal-2024-comparative,SHAHRIAR2025104358}, often using datasets of user-written texts and their associated downstream tasks as proxies for privacy evaluation. While some evaluations demonstrate protection via empirical demonstration of privacy protection on established benchmarks \cite{pilan-etal-2022-text} or against simulated attack vectors \cite{9152761, 299573}, other methods rely on the proof of theoretically grounded privacy guarantees \cite{habernal-2021-differential}. 

Often, however, evaluation of privacy preservation in PPNLP techniques does not take human opinions directly into account, and we were unable to find any evidence of recent works running large-scale human evaluations of privacy in text. One exception comes with a recent work \cite{meisenbacher-etal-2025-investigating} that investigates user perspectives on private texts, albeit with a specific focus on Differential Privacy. This significant gap calls for the greater inclusion of human voices in the evaluation of PPNLP methods, i.e., for measuring whether such methods protect privacy from the perspective of users.

\paragraph{Human-centered (AI) Privacy}
Outside of the realm of NLP, a great deal of work has been performed to better understand the perceptions of humans on what privacy means. Previous works focus on user perceptions of privacy in a number of domains, including social media \cite{doi:10.1177/1461444816660731, baker2022understanding, 10.1145/3555608}, mobile applications \cite{furini2020privacy, 10.1145/3510003.3510079,10.1145/3676529}, e-commerce \cite{bandara2020privacy, ALKIS2022107412}, e-learning \cite{6268044, Chen19052025}, voice-based digital assistants \cite{Ha03042021, vimalkumar2021okay, leschanowsky2024evaluating, 10.1145/3274371}, and other smart home devices \cite{264053, 10.1145/3558095}.

In the context of AI technologies, privacy remains one of the most persistent concerns among users \cite{elliott2022ai}. In a study with over 10,000 people around the world investigating how people expect AI to affect the future, \citet{289526} find that one of the highest perceived negative effects AI will have is that of decreased privacy. Other works support this notion, particularly in the privacy challenges created or exacerbated in the development of AI-based systems \cite{cheng2020ai,wang2024security}. This is especially the case in critical domains such as AI-assisted healthcare \cite{williamson2024balancing}. As such, the privacy issues arising from AI have led privacy to be named a \say{grand challenge} of AI \cite{ozmen2023six}. Yet the concept of privacy itself remains elusive and often misunderstood, making it difficult to determine what exactly people refer to when they express their privacy concerns, and how AI technologies may or may not pose a threat to their interests \cite{elliott2022ai}. 

This lack of clarity in understanding user perceptions of AI and privacy is illustrated by the so-called \textit{privacy paradox}, where despite stated privacy concerns, users might still be driven by perceived usefulness or utility over privacy \cite{willems2023ai,meisenbacher-etal-2025-investigating,zhang2025privacyleakageovershadowedviews}. As stated above, however, it is not always clear what humans actually perceive as private, especially in text data, and therefore, it is difficult to draw conclusions about when users may be prioritizing other factors over privacy.

\paragraph{LLM-as-a-Judge}
Traditional methods for evaluating natural language outputs from language models have been criticized for their lack of consideration of context and requirement for reference texts, such as in the case of BLEU or ROUGE \cite{zhang2019bertscore, liu2023g}. Efforts to improve upon these static, lexical-based methods have extended to semantic similarity-based scores, such as BERTScore \cite{zhang2019bertscore}, yet a semantically similar output may not always correlate with correct outputs.

As a more reliable method than automated metrics, human evaluation has long been held as the gold standard for assessing the generated outputs of language models. Despite the desirable features of human evaluations, running such assessments is costly and time-consuming, thus making them hard to scale \cite{elangovan-etal-2024-considers}. In addition, human raters in evaluation tasks are prone to subjectivity, inconsistency, misunderstandings, and fatigue \cite{bojic2023hierarchical}, thus degrading the reliability of human-based evaluations.

An alternative to human evaluations that has gained popularity alongside the rise of LLMs has been leveraging the so-called \textit{LLM-as-a-Judge} paradigm \cite{li2024llmsasjudgescomprehensivesurveyllmbased, gu2025surveyllmasajudge}, in which LLMs replace humans as the judges in evaluation tasks. This paradigm is rooted in the capability of modern LLMs to recognize and understand the quality and correctness of generated outputs, particularly from being tuned on human preferences. While LLM-as-a-Judge has been shown to contain its own pitfalls \cite{chen-etal-2024-humans, 10.1145/3708359.3712091}, such as being susceptible to \say{beauty bias} and struggling with domain-specific language, several works have shown that LLM-as-a-Judge leads to high alignment with human judgments, often more so than human-human inter-annotator agreement. Thus, the core advantages of LLM-as-a-Judge lie in its ability to consistently and accurately mirror human judgments, while also being scalable and drastically reducing evaluation costs.

Due to the clear benefits of LLM-as-a-Judge for natural language evaluation, we ask whether the same paradigm can be used for assessing the privacy sensitivity of textual data. We are the first to conduct such an investigation, and we seek to provide clarity on the benefits and limitations of LLM-as-a-Judge for privacy evaluation.

\section{Experimental Setup}
Our methodology is divided into four steps, which form the basis for answering our three research questions. First, we follow an adversarial approach to select a representative sample of user-written texts for privacy evaluation. Then, we design two prompts for LLM-as-a-Judge privacy evaluation, proceeding to employ these on 13 LLMs from various model providers. Next, we design and deploy a custom survey application, enabling us to capture the privacy perceptions of human annotators on our selected texts. Finally, we conduct an online survey with 677 participants, leading to an analysis of human-LLM alignment on privacy evaluation in text.

\subsection{Data Selection}
Our goal is to provide both LLMs and human annotators with diverse text samples, so that we can measure privacy perceptions across various domains. To accomplish this, we collected a set of 10 publicly available datasets, all of which are user-written texts and plausibly present some case for private or sensitive data. A text example from each dataset can be found in Table \ref{tab:examples} of the Appendix.

\subsubsection{Datasets}
We briefly introduce all 10 selected datasets.

\textbf{Blog Authorship Corpus (BAC).}
The Blog Authorship Corpus is a collection of nearly 700k blog posts collected from almost 20k authors from blogger.com in 2004. We utilize the subset prepared by \citet{meisenbacher-matthes-2024-thinking}, which contains only the blog posts from the top-10 most frequently occurring authors.

\textbf{Enron Emails (EE).}
Enron Emails is a corpus of roughly 500k email messages collected from the Enron organization, made public in 2003 during the public investigation carried out by the Federal Energy Regulatory Commission in the United States. In particular, we only use emails from the \textit{sent items} folder of each user \cite{meisenbacher-etal-2025-impact}.

\textbf{Medical Questions (MQ).}
The Medical Question Answering dataset\footnote{\url{https://huggingface.co/datasets/Malikeh1375/medical-question-answering-datasets}} contains over one million question-answer pairs resulting from user-written medical questions. We only consider the questions and not the answers.

\textbf{Mental Health Blog (MHB).}
Made available by \citet{boinepelli-etal-2022-leveraging}, the Mental Health Blog dataset contains over 39k posts on a mental health forum between 2011 and 2020, posted from over 9000 unique users. We take the posts from the top-50 writing authors.

\textbf{Reddit Confessions (RC).}
This corpus\footnote{\url{https://huggingface.co/datasets/SocialGrep/one-million-reddit-confessions}} contains one million Reddit posts originating from four \say{confession} subreddits.

\textbf{Reddit Legal Advice (RLA).}
This collection of Reddit posts, introduced by \citet{li-etal-2022-parameter}, contains nearly 100k posts from the \textit{/r/legaladvice} subreddit, and they comprise user-written, informal requests for help on legal matters.

\textbf{Reddit Mental Health Posts (RMHP).}
We utilize a large corpus of over 150k posts from Reddit\footnote{\url{https://huggingface.co/datasets/solomonk/reddit_mental_health_posts}}, taken from five mental health-related subreddits. We only consider the posts from the top-50 most frequently writing authors.

\textbf{Trustpilot Reviews (TR).}
The Trustpilot corpus is a large-scale collection of reviews from the Trustpilot platform, prepared by \citet{10.1145/2736277.2741141}. We utilize the \textit{en-US} section of the corpus, i.e., reviews posted on the American version of the platform. 

\textbf{Twitter (TW).}
We take a sample from the corpus of 100m tweets made available by \citet{enryu_2023_15086029}. In particular, we only consider tweets from the top-100 posting authors.

\textbf{Yelp Reviews (YR).}
We use the subset of Yelp Reviews prepared by \citet{utpala-etal-2023-locally}, which contains nearly 20k user reviews written by 10 unique users from the Yelp platform.

\subsubsection{Constructing an Evaluation Dataset}
Our target was to create a representative sample of 250 texts from the above 10 datasets, thus taking 25 texts from each dataset. For seven of the datasets, the data included information about the author of the dataset, i.e., the ID of the user, or, in the case of Trustpilot, the gender of the author. Using this personal attribute as the basis for an adversarial inference attack, we sought to select a representative sample of 25 texts from each dataset, based on the \say{vulnerability} of these texts to a modeled adversary.

\paragraph{Building an Adversarial Inference Model}
To model the adversary described above, we fine-tuned a lightweight BERT-based language model, namely \textsc{deberta-v3-base} \cite{he2021deberta}, to predict the protected attribute (user ID or gender) given the user text as inputs (review, blog post, etc.). We divided each dataset into an 80/20 train/test split, and fine-tuned a \textsc{deberta-v3-base} model for three epochs on the train split, using the HuggingFace Trainer with all default parameters. Thus, this process yielded seven adversarial classifiers.

For each of the trained adversarial classification models, we used the model for inference on the test splits of the respective datasets, most importantly capturing the confidence score of the predictions. We equate these scores to the \say{vulnerability} of the given texts, or rather, how identifiable the protected attributes are to the adversarial model, with prediction confidence as a proxy. With these prediction scores, we sorted all texts in the test split from least to most vulnerable (lowest to highest adversarial confidence), and selected an evenly split sample of 25 texts along this sorted list. By using this a priori estimator of the sensitivity of texts, we were able to select a representative sample of texts ranging from not sensitive to very sensitive, in the eyes of an adversarial model.

For the three datasets without protected attributes as described above, we took a simple random sample of 25 texts from each dataset. Together, with the 175 texts selected based on adversarial confidence, along with the 75 texts sampled from the remaining three datasets, we formed our base evaluation dataset of 250 texts, which we utilize in the remainder of this work. 

\begin{table}[t!]
\centering
\footnotesize
\caption{Simple prompt for LLM privacy evaluation.}
\begin{tabular}{p{0.95\linewidth}}
\hline
\textbf{Prompt}
\\ \hline
You will be given a user\_text document.\\ 
Your task is to provide a `privacy rating' scoring on how sensitive or private the user\_text is, for example including personal information or indirectly identifiable information. \\
Give your answer as a float on a scale of 1 to 5, where 1 means that the user\_text is not sensitive or private at all, and 5 means that the text contains highly sensitive or private information.\\ 
Do not provide any additional rationale.\\ \\

Provide your feedback as follows:\\ \\ 

Feedback::: \\
Total rating: (your rating, as a float between 1 and 5)\\ \\

Now here is the user text.\\ \\

user\_text: [TEXT GOES HERE] \\\\

Feedback::: \\
Total rating: \\
\hline
\end{tabular}
\label{tab:simple_prompt}
\end{table}

\subsection{LLM-as-a-Judge for Privacy Evaluation}
In order to investigate the feasibility of using LLMs for privacy evaluation, we first modeled the privacy sensitivity in text problem as a Likert-type assessment. We then built two prompts, one simple and one improved, following existing guidelines for LLM-as-a-Judge. Finally, we ran our prompts on selected open and closed LLMs.

\subsubsection{Modeling Privacy}
An important first step for our evaluation process was to model what precisely we were to measure in terms of \textit{privacy}. We decided to model the \textit{privacy sensitivity} of a given text, i.e., how sensitive or private a text is perceived to be (whether by humans or LLMs), based purely on its content. For simplicity, we adopt a five-point Likert scale, with the following defined range of privacy sensitivity and their corresponding descriptions:

\begin{enumerate}
    \item \textbf{Harmless}: completely free of any private or sensitive information, either direct or indirect identifiers.
    \item \textbf{Mostly not private}: may contain some indirect identifiers, but is mostly free of sensitive information.
    \item \textbf{Somewhat private}: contains some direct or indirect identifiers, and can be deemed somewhat personal information.
    \item \textbf{Very private}: contains several direct or indirect identifiers, and it is clear that this text contains personal information.
    \item \textbf{Extremely private}: contains highly sensitive information, such as direct personal identifiers, and the text is clearly something that should not be made public.
\end{enumerate}

\subsubsection{Prompts for Privacy Evaluation}
\label{sec:prompts}
Based on our model of privacy sensitivity in text, we defined two prompts to be used for LLM-as-a-Judge for privacy evaluation. As a baseline, we first designed a simple prompt that does not leverage the above scale, but rather instructs an LLM to provide a score anywhere between 1-5, which mimics a simple approach as proposed in early LLM-as-a-Judge evaluations \cite{10.5555/3666122.3668142}. This prompt is provided in Table \ref{tab:simple_prompt}.

Given that the simple prompt does not provide any guidance to the LLM nor does it operate on a discrete scale, we also create an improved prompt, following best practices for LLM-as-a-Judge evaluation\footnote{\url{https://www.databricks.com/blog/LLM-auto-eval-best-practices-RAG}}. In particular, we included an explicit integer (Likert) scale with descriptions, and we also prompt the LLM to provide reasoning so that the final rating may be more informed. This improved prompt is provided in Table \ref{tab:improved_prompt}.

\begin{table}[t]
\centering
\footnotesize
\caption{Improved prompt for LLM privacy evaluation.}
\begin{tabular}{p{0.95\linewidth}}
\hline
\textbf{Prompt}
\\ \hline
You will be given a user\_text document.\\
Your task is to provide a `privacy rating' scoring on how sensitive or private the user\_text, for example including personal information or indirectly identifiable information.\\
Give your answer on a scale of 1 to 5, where 1 means that the user\_text is not sensitive or private at all, and 5 means that text contains highly sensitive or private information.\\ \\

Here is the scale you should use to build your answer:\\
1: The user\_text is harmless: completely free of any private or sensitive information, either direct or indirect identifiers.\\
2: The user\_text is mostly not private: may contain some indirect identifiers, but is mostly free of sensitive information.\\
3: The user\_text is somewhat private: contains some direct or indirect identifiers, and can be deemed somewhat personal information.\\
4: The user\_text is very private: contains several direct or indirect identifiers, and it is clear that this text contains personal information.\\
5: The user\_text is extremely private: contains highly sensitive information, such as direct personal identifiers, and the text is clearly something that should not be made public.\\ \\

Provide your feedback as follows: \\ \\

Feedback:::\\
Evaluation: (your rationale for the rating, as a text)\\
Total rating: (your rating, as a number between 1 and 5)\\ \\

You MUST provide values for `Evaluation:' and `Total rating:' in your answer. \\ \\

Now here is the user\_text.\\ \\

user\_text: [TEXT GOES HERE] \\ \\

Provide your feedback.\\
Feedback:::\\
Evaluation:\\
\hline
\end{tabular}
\label{tab:improved_prompt}
\end{table}

\subsubsection{Running LLM-as-a-Judge Evaluations}
Using the two designed prompts, we chose a selection of available LLMs as our experimental basis. These included both proprietary and open-source LLMs:
\\ \\
{\small
\noindent \underline{Proprietary}:

\noindent\textbf{OpenAI}: \textsc{gpt-4o-mini} (2024-07), \textsc{gpt-4o} (2024-08), \textsc{gpt-4.1} (2025-04) \cite{openai2024gpt4ocard, openai2024gpt4technicalreport}

\noindent\textbf{Google}: \textsc{gemini-2.0-flash}, \textsc{gemini-2.5-flash-preview-05-20} \cite{geminiteam2025geminifamilyhighlycapable}
\\ \textbf{Anthropic}: \textsc{claude-3-5-haiku-20241022}, \textsc{claude-3-7-sonnet-20250219} \footnote{\url{https://www-cdn.anthropic.com/de8ba9b01c9ab7cbabf5c33b80b7bbc618857627/Model_Card_Claude_3.pdf}}

\vspace{10pt}
\noindent \underline{Open-source}:

\noindent\textbf{Meta}: \textsc{Llama-3.2-1B-Instruct}, \textsc{Llama-3.2-3B-Instruct}, \textsc{Llama-3.3-70B-Instruct-Turbo}\footnote{Note that the 70B variant is the only open-source model we did not run locally but instead used \url{https://www.together.ai/}.} \cite{grattafiori2024llama3herdmodels}

\noindent \textbf{Google}: \textsc{gemma-3-1b-it}, \textsc{gemma-3-4b-it}, \textsc{gemma-3-12b-it} \cite{gemmateam2024gemmaopenmodelsbased}
}
\\ \\
For each of the two prompts and 13 selected LLMs, we ran a user text five times through each model, for each of the selected 250 texts. The five scores for each text were averaged for a final score. In the case that a score was not parseable from the output text, this run was simply discarded and not included in the average. For the improved prompt, we also stored the \say{feedback} text outputted by each model, to capture the reasoning behind a particular rating.

\subsection{Survey Design}
We built a custom survey application to gauge privacy perceptions of humans regarding the sensitivity of our selected user texts. We sought to mimic the prompting approach taken with the LLMs, in that the survey participants would receive identical instructions as with the LLMs. The design and makeup of our survey application are described in the following.

\subsubsection{Survey Architecture}
\begin{figure*}[htbp]
    \centering
    \includegraphics[width=0.85\linewidth]{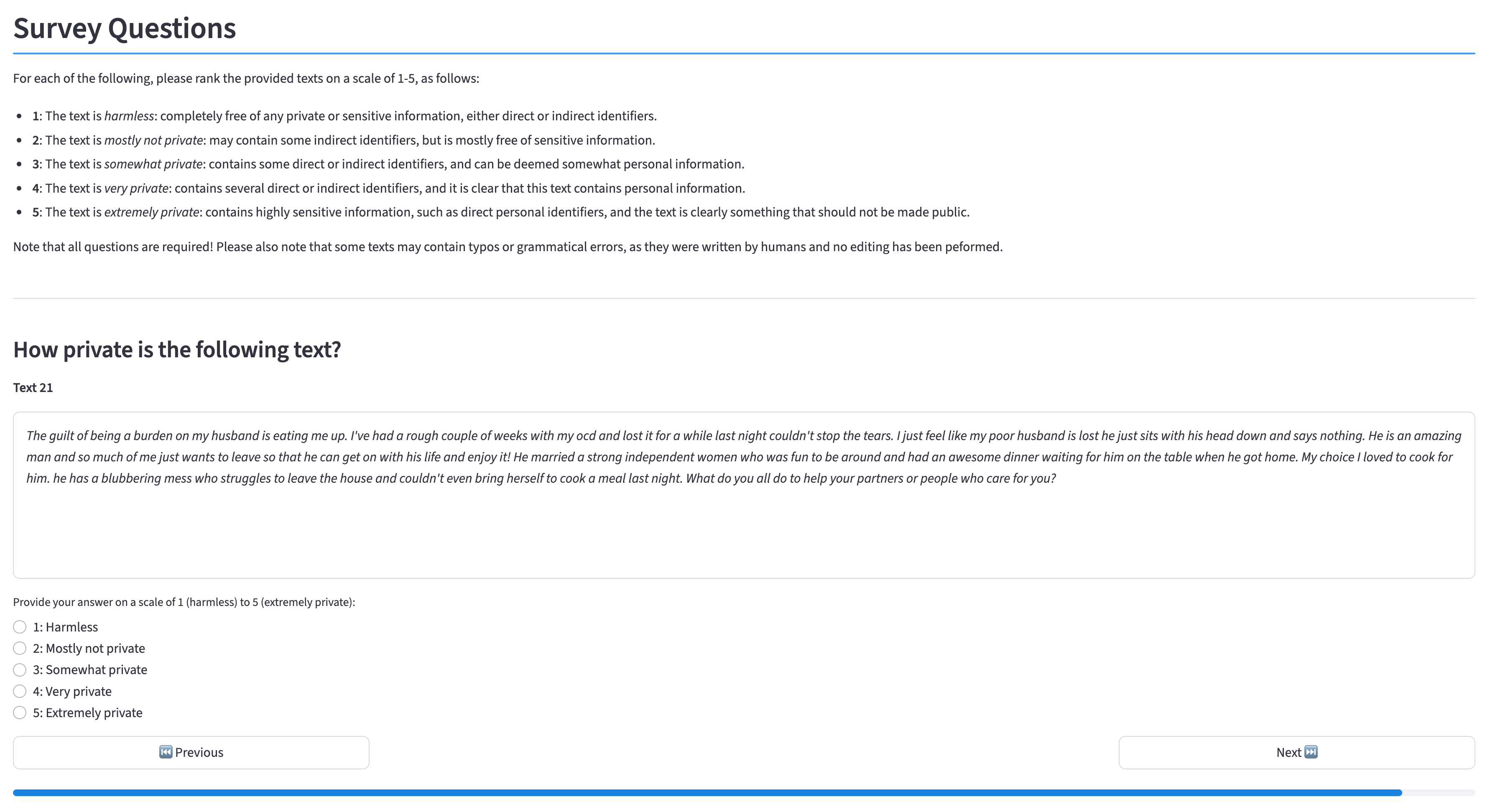}
    \caption{The main survey stage, where participants were prompted to rate each presented text on a privacy scale of 1-5.}
    \label{fig:app}
\end{figure*}

The custom survey application was built using Streamlit\footnote{\url{https://streamlit.io/}} and was deployed publicly on the Streamlit Community Cloud\footnote{The survey application is publicly available at \url{https://privacy-in-text.streamlit.app/}}. All survey responses were stored in a Cloud MongoDB for later analysis. The survey flow was divided into the following sections (pages):

\begin{enumerate}
    \item \textbf{Welcome and Instructions}: We informed participants about the estimated duration, provided the privacy scale as in Table \ref{tab:improved_prompt}, made it clear that privacy is subjective and there are no wrong answers, and finally included a disclaimer that some text examples may contain emotional or vulgar language.
    \item \textbf{Background}: We asked general demographic questions, including country of origin, age (by range), gender (male, female, non-binary, prefer not to answer), highest attained education level, and current employment status.
    \item \textbf{Privacy Rating}: We presented a page with the primary survey questions, which once again provided the privacy scale and prompted the user to rate each text. A progress bar was included to inform the participant of their progress. A screenshot of this primary survey stage is given in Figure \ref{fig:app}.
    \item \textbf{Freeform Feedback}: Following the successful answering of all presented texts, the participants were prompted with the opportunity to answer one (optional) freeform question: 
        \begin{quote}
            \textit{What were your reasoning patterns behind answering the previous questions? In other words, how did you decide the degree to which texts seemed private to you?}
        \end{quote}
   \item \textbf{Submission}: The participants were prompted to submit their responses, thus concluding the survey.
\end{enumerate}

\subsubsection{Survey Question Selection}
Since our selected dataset of 250 user texts was not feasible to be annotated by each survey participant, we conducted a between-subjects survey, where each survey participant would be randomly assigned 20 texts from the pool of 250. This number of 20 texts was decided upon after initial pilot tests within our research group, which indicated that roughly 20 answers could be comfortably given within our 15-minute target time. With this quantity per survey participant, we set a goal that each text was to be rated by at least 50 different participants, thus requiring at least 12,500 total survey answers. At 20 answers per participant, this required a minimum of 625 survey respondents. 

In order to ensure the correct random assignment of questions, we built a mechanism into the backend of the survey, which randomly selected two user texts per individual dataset (thus, 20 in total). This random selection, though, was only performed among the $n$ texts per dataset that currently had the least total responses. For example, if five texts from the Enron dataset had not yet been seen by any participant, while the other 20 had, then a random selection was performed among these five choices. If there was only one minimum, then only the second choice would be randomized, unless, of course, the second lowest also was a singleton.

\subsubsection{Pilot Testing}
A pilot test was run with 10 participants, made up of close colleagues, family, and friends. The goal of this pilot was to identify technical errors, points of ambiguity, or typos. As a result of the pilots, only minor typos in the instructions or questions were found, which were promptly fixed. The responses of the 10 pilot participants are included in the final analysis.

\subsection{Survey Conduction}
We used Prolific\footnote{\url{https://www.prolific.com/}} for survey participant recruitment. We prescreened participants via Prolific according to a number of criteria, including self-reported fluency in English, at least a high school degree, a Prolific approval rating of at least 95\%, and at least 50 previous survey studies completed. We also set the minimum age to 21, due to the potentially sensitive topics covered in our presented texts. We did not limit the survey to any geographic region.

Based on the pilots, we set the estimated survey duration to 15 minutes. We offered \textsterling2.25 per valid submission, at a rate of \textsterling9/hour (marked \say{good} by Prolific). We opted to recruit 675 participants (50 more than our minimum) to provide a buffer. After the recruitment of all 675 participants, the median completion time was 17.4 minutes, resulting in a final hourly wage of  \textsterling7.75 (marked \say{fair} by Prolific).

As required by Prolific, we incorporated two \textit{attention checks} into the survey, which took the form of a privacy rating question (Figure \ref{fig:app}), however, with the text replaced as an explicit instruction to select \say{3: Somewhat private}. As Prolific requires both attention checks to be failed for a rejected submission, no submissions were rejected (i.e., all participants were compensated). However, for our analysis, we screened out eight submissions where at least one check was failed, resulting in a final pool of 677 valid submissions.

\subsubsection{Survey Participant Demographics}
Comprising the 677 valid submissions were participants from 45 countries spanning all six major continents. The most came from Europe (n=287, 42.3\%), followed by Africa (n=236, 34.9\%), North America (n=69, 10.2\%), Asia (n=35, 5.2\%), South America (n=26, 3.8\%), and Oceania (n=24, 3.5\%).

The slight majority of the participants identified as male (n=366, 54.1\%), with females also strongly represented (n=302, 44.6\%). 8 participants identified as non-binary, and one preferred not to answer. The most participants were between the ages of 25-34 (n=297, 43.9\%), followed by 35-44 (n=137, 20.2\%), 21-24 (n=109, 16.1\%), 45-55 (n=79, 11.7\%), and 55+ (n=55, 8.1\%). Nearly half of the participants held a Bachelor's degree (n=324, 47.9\%); others reported holding a Master's degree (n=182, 26.9\%), High School degree (n=96, 14.2\%), Doctorate (n=51, 7.5\%), or Associate's degree (n=24, 3.5\%).

\section{Results}
We structure the presentation of our results by our three research questions, first looking at inter-LLM agreement in our privacy evaluations, then proceeding to inter-human agreement. Finally, we analyze the alignment of LLMs and humans in the task of privacy evaluation, including a breakdown by dataset and demographic.

\begin{figure*}[htbp]
    \centering
    \begin{subfigure}{0.49\textwidth}
        \centering
        \includegraphics[width=\linewidth]{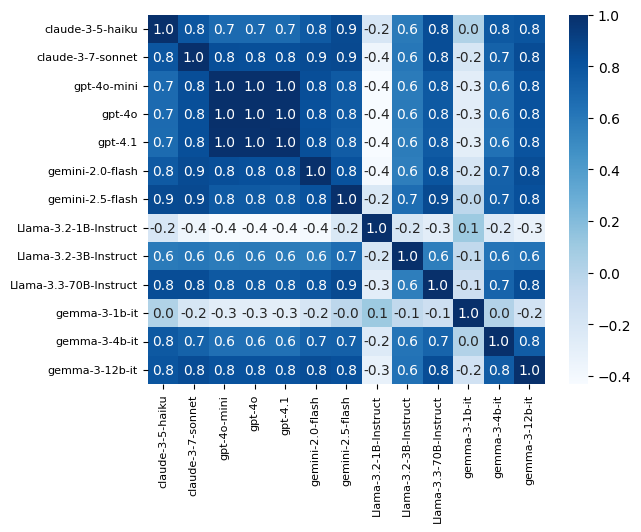}
        \caption{\textit{Simple} Prompt (Overall Agreement = 0.54)}
        \label{fig:llm_simple_agreement}
    \end{subfigure}
    \hfill
    \begin{subfigure}{0.49\textwidth}
        \centering
        \includegraphics[width=\linewidth]{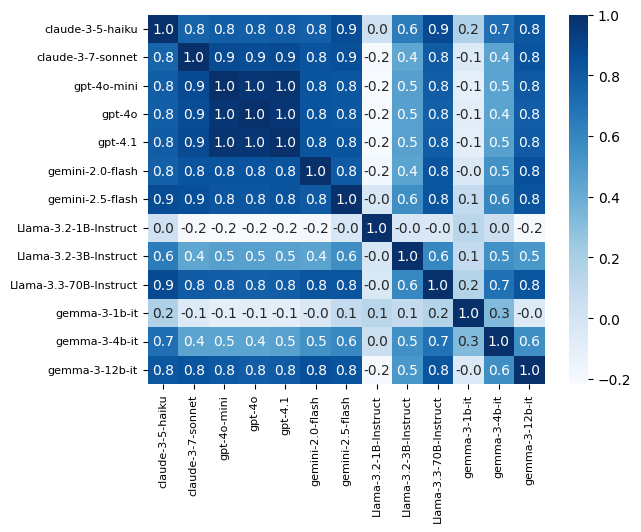}
        \caption{\textit{Improved} Prompt (Overall Agreement = 0.58)}
        \label{fig:llm_imrpoved_agreement}
    \end{subfigure}
    \caption{LLM Agreement (Krippendorff's alpha) across all utilized LLMs and all 250 annotated texts.}
    \label{fig:llm_agreement}
\end{figure*}

\subsection{RQ1: LLM-LLM Agreement on Privacy}
To measure the levels of agreement between our 13 selected LLM annotators, we employ the \textit{Krippendorff's alpha} inter-rater reliability statistic. The choice of this statistic is primarily based on the results of the conducted survey with humans, in which each annotator only rates 20 texts of the total 250. Krippendorff's alpha is capable of handling such missing values, as it only measures the extent of agreement between raters \textit{on the same units of analysis}, or the same texts scored between raters. An alpha score of 1 indicates perfect agreement, 0 implies roughly the same performance as random chance, and -1 indicates much poorer performance than random chance. Thus, for the remainder of this work, we equate \textit{agreement} with the associated Krippendorff's alpha scores. For all agreement calculations, we employ the  \textsc{krippendorff} Python library \cite{castro-2017-fast-krippendorff}.

\subsubsection{Simple vs. Improved Prompt}
In Section \ref{sec:prompts}, we introduced the two prompts we use for privacy evaluation with the 13 selected LLMs. As a first point of analysis, we measure the overall agreement between the 13 LLM raters, both for the \textit{simple} and \textit{improved} prompts. We note that for this and all ensuing agreement calculations, we round the average ratings (i.e., the average between fives per text) to the nearest integer, thus enabling a fair comparison on the five-point Likert scale we define. With these, we calculated an overall inter-LLM agreement of \textbf{0.54} for the simple prompt, and \textbf{0.58} for the improved prompt. Although the improved prompt does indeed allow for a more guided privacy decision, both agreement scores indicate generally lower inter-LLM agreement \cite{marzi2024k}.

Diving deeper, we calculate the pairwise agreement between all LLM annotator pairs over all 250 texts. This paints a more precise picture of which individual LLMs are more aligned with each other, and which may be decreasing the overall agreement. The results of this analysis, for both prompting strategies, are found in Figure \ref{fig:llm_agreement}.

We see that the shift from the simple to improved prompt generally increases alignment; however, this seems to be most effective in aligning proprietary LLMs with each other (e.g., GPT models with Claude models), and a similar effect can be observed between the open-source models. Interestingly, model size plays an important role, as only the large open-source models (\textsc{Llama-3.3-70B} and \textsc{gemma-3-12b}) are highly in agreement with the proprietary LLMs.

Due to the clear importance of model choice, we extend our investigation of inter-LLM agreement, isolating the factors of model availability (open vs. closed), model family (i.e., model provider), and model size. Within \textit{closed} LLMs (\textsc{GPT}, \textsc{Claude}, and \textsc{Gemini}), there is a high agreement score of \textbf{0.84}, whereas open models (\textsc{Llama} and \textsc{gemma}) achieve a much lower score of \textbf{0.37}. The three OpenAI models achieve the highest agreement (\textbf{0.98}), whereas the three \textsc{gemma} models achieve the lowest agreement (0.58), with \textsc{Claude} (\textbf{0.86}), \textsc{Gemini} (\textbf{0.84}), and \textsc{Llama} (\textbf{0.63}) in between. As all inter-family agreement scores are at or above the overall agreement, this implies that models are more cohesive with their privacy reasoning within a family as opposed to with models from other providers.

Finally, focusing on model size, we study only the utilized open-source models, due to the generally unknown size of the proprietary models. For simplicity, we calculate the agreement between the \textit{large} models (\textsc{Llama-3.3-70B} and \textsc{gemma-3-1b}), \textit{medium} models (\textsc{Llama-3.2-3B} and \textsc{gemma-3-4b}). and \textit{small} models (\textsc{Llama-3.2-1B} and \textsc{gemma-3-1b}), which achieve agreement scores of \textbf{0.83}, \textbf{0.55}, and \textbf{0.13}, respectively. These results suggest that model size matters significantly when using LLM as privacy evaluators, with larger models providing much more reliable and consistent ratings.

For the remainder of this work, particularly in comparing LLM to human responses, we use solely the LLM ratings resulting from the \textit{improved} prompt, due to the generally higher agreement scores, as well as to align with the human annotators in the survey, who were provided the improved prompt as instructions.

\subsection{RQ2: Human Perceptions of Privacy in Text}
We now present and analyze the results of the conducted survey with 677 human participants. Each text on our full dataset was scored by an average of 55.2 raters, and the overall agreement score between annotators is \textbf{0.39}. This low agreement gives an initial insight into the differing privacy views between individuals, especially when taking into account diverse demographic factors.

While the overall agreement score is quite low, we also calculate the \textit{average pairwise} agreement between all human annotators. This yields an agreement score of \textbf{0.54}, which indicates that, on average, any two human annotators will agree with each other to a greater degree than when viewing the whole annotator sample.

Observing the individual sub-demographics, we find that those from North America exhibit the greatest overall agreement (\textbf{0.49}), followed by Europe (\textbf{0.45}), Oceania (\textbf{0.44}), South America (\textbf{0.41}), Asia (\textbf{0.39}), and Africa (\textbf{0.31}). While we caution that some of the less represented continent agreement scores may not be as reliable, there do exist considerable agreement gaps between fairly equally represented demographics, such as between Europe and Africa.

Looking to the factor of age, the highest agreement was observed with 21-24 year olds (\textbf{0.41}), the lowest with 35-44 (\textbf{0.36}), and the other age groups all had a rounded agreement of \textbf{0.40}. Male participants had higher agreement (\textbf{0.42}) than female participants (\textbf{0.37}). Interestingly, participants with a Doctorate showed very low levels of agreement (\textbf{0.22}), followed by Master's and Bachelor's (both \textbf{0.40}), High School (\textbf{0.43}), and Associate's (\textbf{0.44}). Thus, we see a generally decreasing agreement score as education level increases.

\subsection{RQ3: Human-LLM Alignment on Privacy}
Now that we have analyzed both inter-LLM and inter-human alignment of privacy evaluation in text, we proceed to explore the \textit{alignment} between the two, namely, to what degree LLMs represent the general privacy attitudes of humans regarding textual data.

\begin{figure}[t]
    \centering
    \includegraphics[scale=0.45]{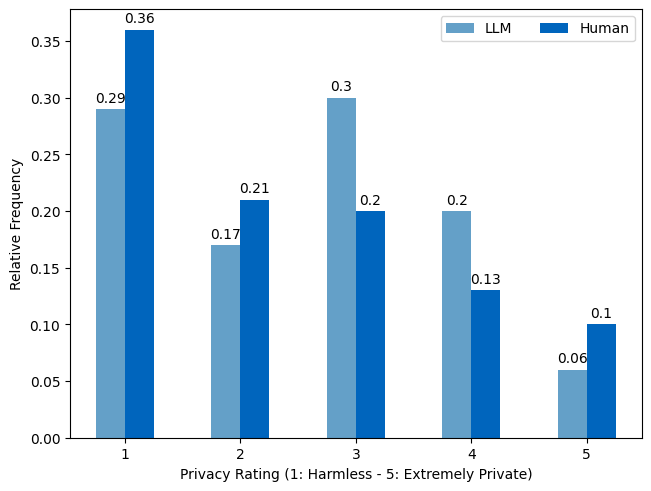}
    \caption{Relative frequencies of the five privacy selection choices over all rated texts, both for the 13 LLM annotators (``LLM'') and the 677 survey participants (``Human'').}
    \label{fig:freq}
\end{figure}

We begin our investigation with a broad comparison of the \textit{selection choices} between LLMs and humans. In Figure \ref{fig:freq}, we depict the relative frequencies of the five privacy levels, i.e., the number of times a privacy rating (1-5) was selected, divided by the total number of votes (LLM experiments or survey responses). The results of this analysis show that humans, on average, choose lower values (1 and 2), while LLMs tend to score the privacy sensitivity of texts as higher (3 and 4). Also notable is the decreasing frequency of response selections for humans as privacy level increases, while LLM choices are more centered around 3 (somewhat private).

\begin{figure}[t]
    \centering
    \includegraphics[scale=0.55]{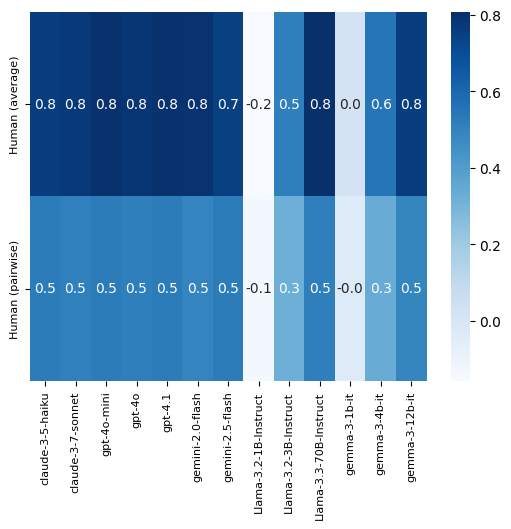}
    \caption{Agreement scores between each LLM and the average human rating (top), compared to the average pairwise agreement between LLMs and humans (bottom).}
    \label{fig:human_llm}
\end{figure}

In investigating further the alignment between LLMs and humans in our privacy evaluation task, we distinguish between two entities: agreement with the \textit{average} (global) human rating per text and the \textit{pairwise} agreement between an LLM and each survey participant. Figure \ref{fig:human_llm} illustrates the differences in measuring these two entities, in that while LLMs show very strong agreement with the average human ratings, pairwise agreement indicates lower agreement. This result is reasonable, as LLMs may very accurately capture the global notion of privacy sensitivity in textual data, but cannot easily mirror differing individual human opinions. 

Building on these findings, we conduct a final extensive analysis, in which we compare the average ratings of all selected LLMs to different subsets of human annotators, based on our captured demographics. The results of this analysis are presented in Table \ref{tab:results}. Globally speaking, we observe that LLMs, on average, effectively capture the average ratings of humans across all datasets. Furthermore, we can find in the case of every dataset an LLM that exactly or with a difference of $\pm 0.1$ resembles the average human privacy rating. We also note that despite fluctuations, the \textit{average} ratings between demographic groups remain quite steady, a result that extends beyond simple agreement scores.

\begin{table*}[htbp]
\small
\caption{Average privacy ratings across all 10 datasets, broken down by LLM annotators and human demographics. The scores represent the average privacy rating for the 25 selected texts in each dataset, with the subscript denoting the standard deviation. Bolded LLM scores represent the closest result to the global human average, with ties being broken by the smallest standard deviation. The score of `-' results from no parseable answers being extracted from all runs on the texts of the MQ dataset.}
\begin{tabular}{l|cccccccccc|c}
\multicolumn{1}{r|}{Dataset:} & BAC & EE & MQ & MHB & RC & RLA & RMHP & TR & TW & YR & Average \\ \hline
\textbf{LLM (global)} & $2.1_{1.0}$ & $2.6_{1.1}$ & $2.7_{1.4}$ & $3.8_{0.7}$ & $3.5_{1.1}$ & $3.1_{0.8}$ & $2.9_{1.0}$ & $1.7_{0.9}$ & $1.6_{0.9}$ & $1.9_{1.0}$ & $2.6_{1.2}$ \\ \hline
\textsc{gpt-4o-mini} &  $1.4_{0.7}$ & $2.2_{1.2}$ & $2.5_{1.3}$ & $3.4_{0.6}$ & $\mathbf{3.2_{0.9}}$ & $\mathbf{2.9_{0.6}}$ & $2.4_{0.9}$ & $1.2_{0.4}$ & $1.0_{0.0}$ & $1.3_{0.4}$ & $2.2_{1.2}$ \\
\textsc{gpt-4o} & $1.4_{0.7}$ & $2.2_{1.2}$ & $2.5_{1.4}$ & $3.4_{0.6}$ & $\mathbf{3.2_{0.9}}$ & $3.0_{0.6}$ & $2.4_{0.8}$ & $1.2_{0.5}$ & $1.0_{0.0}$ & $1.4_{0.5}$ & $2.2_{1.2}$ \\
\textsc{gpt-4.1} & $1.4_{0.7}$ & $2.2_{1.2}$ & $2.5_{1.3}$ & $3.5_{0.6}$ & $\mathbf{3.2_{0.9}}$ & $3.0_{0.6}$ & $2.4_{1.0}$ & $1.2_{0.4}$ & $1.0_{0.0}$ & $1.4_{0.5}$ & $2.2_{1.2}$ \\
\textsc{gemini-2.0-flash} & $1.9_{0.9}$ & $2.4_{0.9}$ & $2.4_{1.3}$ & $\mathbf{3.2_{0.6}}$ & $3.2_{1.0}$ & $3.0_{0.6}$ & $2.7_{0.8}$ & $\mathbf{1.4_{0.6}}$ & $1.1_{0.3}$ & $\mathbf{1.6_{0.7}}$ & $\mathbf{2.3_{1.1}}$  \\
\textsc{gemini-2.5-flash-preview} &  $1.8_{0.9}$ & $2.5_{1.1}$ & $2.7_{1.7}$ & $4.3_{0.8}$ & $3.7_{1.2}$ & $3.0_{0.9}$ & $2.9_{1.3}$ & $1.3_{0.6}$ & $1.1_{0.3}$ & $1.4_{0.7}$ & $2.5_{1.5}$  \\
\textsc{claude-3-5-haiku} &  $2.0_{0.9}$ & $2.9_{1.2}$ & $2.9_{1.7}$ & $4.5_{0.6}$ & $4.0_{1.2}$ & $3.4_{0.9}$ & $3.2_{1.0}$ & $1.4_{0.7}$ & $\mathbf{1.6_{0.7}}$ & $1.4_{0.7}$ & $2.7_{1.4}$ \\
\textsc{claude-3-7-sonnet} &  $1.6_{0.7}$ & $2.0_{0.9}$ & $2.4_{1.4}$ & $3.5_{0.7}$ & $3.2_{1.2}$ & $2.6_{0.6}$ & $\mathbf{2.6_{1.1}}$ & $1.2_{0.4}$ & $1.0_{0.2}$ & $1.4_{0.6}$ & $2.1_{1.2}$\\
\textsc{Llama-3.2-1B-Instruct} &  $3.5_{0.6}$ & $3.6_{0.6}$ & $3.7_{0.8}$ & $3.8_{0.5}$ & $4.3_{0.7}$ & $3.5_{0.7}$ & $3.5_{0.8}$ & $3.2_{0.8}$ & $3.2_{0.8}$ & $3.7_{0.8}$ & $3.6_{0.8}$  \\
\textsc{Llama-3.2-3B-Instruct} &  $2.3_{0.7}$ & $2.9_{1.1}$ & $\mathbf{2.7_{1.1}}$ & $3.9_{0.8}$ & $3.8_{0.8}$ & $3.5_{0.8}$ & $3.2_{1.1}$ & $2.2_{0.9}$ & $2.4_{1.2}$ & $1.9_{1.0}$ & $2.9_{1.2}$  \\
\textsc{Llama-3.3-70B-Instruct} & $\mathbf{2.2_{0.8}}$ & $2.7_{1.0}$ & $2.5_{1.4}$ & $3.8_{0.8}$ & $3.8_{1.1}$ & $3.1_{0.5}$ & $3.0_{0.8}$ & $1.6_{0.7}$ & $1.3_{0.5}$ & $\mathbf{1.8_{0.7}}$ & $2.6_{1.2}$  \\
\textsc{gemma-3-1b-it} & $2.9_{0.8}$ & $2.9_{0.5}$ & -- & $3.8_{0.8}$ & $3.6_{0.9}$ & $3.5_{0.5}$ & $3.2_{0.5}$ & $2.9_{0.5}$ & $2.9_{0.7}$ & $3.4_{0.6}$ & $3.2_{0.7}$  \\
\textsc{gemma-3-4b-it} & $2.6_{1.1}$ & $3.4_{0.9}$ & $3.0_{1.2}$ & $4.0_{0.0}$ & $3.5_{1.0}$ & $3.8_{0.5}$ & $3.2_{0.8}$ & $2.5_{0.8}$ & $2.0_{1.0}$ & $2.6_{0.7}$ & $3.0_{1.0}$  \\
\textsc{gemma-3-12b-it} & $1.8_{0.8}$ & $\mathbf{2.5_{0.9}}$ & $2.4_{1.4}$ & $3.6_{0.5}$ & $3.0_{1.2}$ & $2.7_{1.0}$ & $2.7_{0.9}$ & $1.3_{0.5}$ & $1.2_{0.5}$ & $1.4_{0.7}$ & $2.2_{1.2}$  \\ \hline \hline
\textbf{Human (global)} & $2.1_{1.1}$ & $2.5_{1.4}$ & $2.6_{1.4}$ & $3.2_{1.3}$ & $3.3_{1.4}$ & $2.9_{1.2}$ & $2.6_{1.3}$ & $1.4_{0.8}$ & $1.6_{1.0}$ & $1.7_{0.9}$ & $2.4_{1.4}$
  \\ \hline
Africa (n=236) &  $2.0_{1.2}$ & $2.5_{1.4}$ & $2.8_{1.5}$ & $3.0_{1.3}$ & $3.3_{1.3}$ & $3.0_{1.3}$ & $2.7_{1.3}$ & $1.6_{0.9}$ & $1.7_{1.1}$ & $1.8_{1.0}$ & $2.4_{1.4}$ \\
Asia (n=35) & $2.1_{1.1}$ & $2.5_{1.3}$ & $2.8_{1.4}$ & $3.3_{1.2}$ & $3.2_{1.3}$ & $2.7_{1.1}$ & $2.6_{1.3}$ & $1.4_{0.7}$ & $1.4_{0.8}$ & $2.0_{1.0}$ & $2.4_{1.3}$ \\
Europe (n=287) & $2.1_{1.1}$ & $2.6_{1.4}$ & $2.6_{1.4}$ & $3.2_{1.2}$ & $3.3_{1.4}$ & $2.8_{1.1}$ & $2.6_{1.2}$ & $1.4_{0.7}$ & $1.6_{1.0}$ & $1.7_{0.9}$ & $2.4_{1.3}$  \\
North America (n=69) & $1.9_{1.1}$ & $2.4_{1.3}$ & $2.5_{1.4}$ & $3.5_{1.3}$ & $3.4_{1.4}$ & $2.9_{1.2}$ & $2.6_{1.3}$ & $1.4_{0.6}$ & $1.6_{1.0}$ & $1.6_{0.9}$ & $2.4_{1.4}$ \\
Oceania (n=24) & $1.9_{1.0}$ & $2.6_{1.4}$ & $2.5_{1.1}$ & $3.0_{1.2}$ & $3.2_{1.3}$ & $2.6_{1.1}$ & $2.2_{1.1}$ & $1.2_{0.8}$ & $1.4_{0.8}$ & $1.6_{0.8}$ & $2.2_{1.3}$  \\
South America (n=26) & $2.2_{1.1}$ & $2.2_{1.4}$ & $2.1_{1.2}$ & $3.1_{1.3}$ & $2.9_{1.2}$ & $2.8_{1.0}$ & $2.6_{1.3}$ & $1.3_{0.6}$ & $1.3_{0.7}$ & $1.6_{0.8}$ & $2.2_{1.2}$ \\ \hline
18-24 (n=109) & $2.1_{1.2}$ & $2.5_{1.4}$ & $2.7_{1.5}$ & $3.1_{1.2}$ & $3.3_{1.3}$ & $2.9_{1.2}$ & $2.6_{1.3}$ & $1.4_{0.8}$ & $1.6_{1.0}$ & $1.8_{1.0}$ & $2.4_{1.3}$  \\
25-34 (n=297) & $2.1_{1.1}$ & $2.6_{1.4}$ & $2.7_{1.5}$ & $3.3_{1.3}$ & $3.3_{1.4}$ & $3.0_{1.2}$ & $2.7_{1.3}$ & $1.5_{0.8}$ & $1.6_{1.0}$ & $1.8_{1.0}$ & $2.4_{1.4}$ \\
35-44 (n=137) & $2.0_{1.2}$ & $2.5_{1.4}$ & $2.5_{1.4}$ & $3.0_{1.3}$ & $3.2_{1.4}$ & $2.7_{1.2}$ & $2.5_{1.2}$ & $1.4_{0.8}$ & $1.6_{1.1}$ & $1.7_{0.9}$ & $2.3_{1.3}$ \\
45-55 (n=79) & $2.0_{1.1}$ & $2.4_{1.4}$ & $2.6_{1.4}$ & $3.2_{1.2}$ & $3.2_{1.3}$ & $2.9_{1.1}$ & $2.7_{1.3}$ & $1.4_{0.9}$ & $1.6_{0.9}$ & $1.7_{0.9}$ & $2.4_{1.3}$\\
55+ (n=55) & $2.1_{1.2}$ & $2.5_{1.4}$ & $2.6_{1.4}$ & $3.2_{1.4}$ & $3.4_{1.5}$ & $2.7_{1.2}$ & $2.5_{1.3}$ & $1.4_{0.7}$ & $1.8_{1.2}$ & $1.6_{0.8}$ & $2.4_{1.4}$ \\ \hline
Female (n=302) & $2.0_{1.2}$ & $2.5_{1.4}$ & $2.6_{1.4}$ & $3.1_{1.3}$ & $3.3_{1.4}$ & $2.9_{1.2}$ & $2.6_{1.3}$ & $1.5_{0.8}$ & $1.7_{1.1}$ & $1.7_{1.0}$ & $2.4_{1.4}$ \\
Male (n=366) & $2.1_{1.1}$ & $2.5_{1.4}$ & $2.7_{1.4}$ & $3.2_{1.3}$ & $3.3_{1.4}$ & $2.9_{1.2}$ & $2.7_{1.3}$ & $1.4_{0.7}$ & $1.6_{1.0}$ & $1.8_{0.9}$ & $2.4_{1.3}$ \\
Non-binary (n=8) & $2.3_{1.5}$ & $3.1_{1.7}$ & $3.1_{1.6}$ & $4.0_{0.9}$ & $3.6_{1.0}$ & $3.4_{0.9}$ & $2.8_{1.4}$ & $2.2_{1.4}$ & $1.8_{1.3}$ & $2.4_{1.5}$ & $2.9_{1.5}$ \\ \hline
High School (n=96) & $2.0_{1.2}$ & $2.5_{1.4}$ & $2.5_{1.4}$ & $3.1_{1.3}$ & $3.3_{1.4}$ & $2.8_{1.2}$ & $2.4_{1.2}$ & $1.3_{0.7}$ & $1.5_{1.0}$ & $1.6_{0.8}$ & $2.3_{1.4}$ \\
Associate's or equivalent (n=24) & $1.9_{1.0}$ & $2.4_{1.4}$ & $2.5_{1.4}$ & $3.2_{1.5}$ & $3.1_{1.4}$ & $2.8_{1.2}$ & $2.6_{1.2}$ & $1.5_{0.7}$ & $1.5_{0.9}$ & $1.6_{0.8}$ & $2.3_{1.3}$ \\
Bachelor's or equivalent (n=324) & $2.0_{1.1}$ & $2.5_{1.4}$ & $2.6_{1.5}$ & $3.2_{1.2}$ & $3.2_{1.3}$ & $2.8_{1.2}$ & $2.6_{1.2}$ & $1.4_{0.8}$ & $1.6_{1.0}$ & $1.8_{0.9}$ & $2.4_{1.3}$ \\
Master's or equivalent (n=182) & $2.1_{1.1}$ & $2.6_{1.4}$ & $2.8_{1.5}$ & $3.3_{1.3}$ & $3.3_{1.4}$ & $3.0_{1.3}$ & $2.6_{1.3}$ & $1.5_{0.8}$ & $1.6_{1.0}$ & $1.7_{0.9}$ & $2.4_{1.4}$\\
Doctorate or equivalent (n=51) & $2.2_{1.2}$ & $2.6_{1.4}$ & $2.8_{1.3}$ & $3.2_{1.4}$ & $3.2_{1.3}$ & $3.2_{1.2}$ & $2.8_{1.3}$ & $1.6_{1.0}$ & $1.8_{1.3}$ & $2.0_{1.1}$ & $2.5_{1.4}$
\end{tabular}
\label{tab:results}
\end{table*}

\section{Discussion}
In the following, we reflect on the main findings of our work, and discuss their implications for the field of privacy-preserving NLP.

\subsection{Are humans and LLMs aligned on privacy?}
Our experiment results tell an interesting story, both in the inter-LLM and inter-human agreement scores, as well as in the alignment between LLMs and humans. We find that LLMs offer a reliable and consistent method for assessing the privacy sensitivity of texts, as indicated by the fact that many of the pairwise agreement scores between LLMs are higher than the average pairwise human agreement (0.54). These results provide evidence that, in the very subjective notion of \textit{privacy}, particularly as we define it in our work, LLMs may generally be a more objective evaluator than humans. 

From an LLM point of view, however, we find that the choice of LLM has a significant impact on the resulting privacy score coherence. Generally, larger, proprietary LLMs provide the best results in terms of coherence and agreement with the human privacy ratings, and any \say{smaller} open-source LLMs struggle to achieve credible results. A potential compromise between performance and the ability to run evaluations locally comes with the larger open-source models we tested (\textsc{Llama-3.3-70B} and \textsc{gemma-3-12b}), which achieve nearly comparable performance to the closed models. Regardless, though, we show that effective LLM-as-a-Judge privacy evaluations require capable models with higher amounts of instilled knowledge, requiring both financial and hardware resources.

When viewing the results presented in both Figure \ref{fig:freq} and Table \ref{tab:results}, we show that, on average, LLMs may tend to \textit{overestimate} the privacy sensitivity of texts as compared to humans, as demonstrated by the often higher average privacy ratings, both per dataset and globally. This is apparent in the quite large gaps in relative frequencies in privacy scores, shown in Figure \ref{fig:freq}, where humans prefer 1 (harmless) and 2 (mostly not private), and LLMs select 3 (somewhat private) and 4 (very private) more often. We explore possible reasons behind this finding in the next discussion point.

An important finding in the discussion of human-LLM alignment on privacy in text data becomes \textit{who} is being represented when we measure agreement. Figure \ref{fig:human_llm} makes the important point that if the goal is to represent the \say{global human privacy opinion}, i.e., the average human rating, then LLMs achieve very high agreement. This is less so the case when considering average pairwise agreement between an LLM annotator and each human annotator, which achieves comparable results to average human-human pairwise agreement. Thus, we find that LLMs are aligned well on the \textit{general} perception of privacy, whereas they cannot capture the unique perspectives and experiences of all represented demographics.

\subsection{\textit{Sensitive} vs. \textit{Identifiable}: Drawing from human and LLM reasoning patterns}
In light of the high general agreement between LLMs and humans, but at the same time with a slight tendency for LLMs to rate a text as more private, we turn to the reasons provided by both parties for their privacy ratings. This is enabled by the feedback option we provided in our survey, as well as by capturing the \say{feedback} section of the response as required in our improved LLM prompt.

From the 677 survey submissions, we received 479 responses to the feedback question. To analyze these, we first used Sentence Transformers \cite{reimers-gurevych-2019-sentence}, specifically using \textsc{jinaai/jina-embeddings-v3} \cite{sturua2024jinaembeddingsv3multilingualembeddingstask}, to embed each response, and then we used Agglomerative Clustering (\textsc{sklearn}) to cluster the responses by semantic similarity. As a team, we manually verified these clusters, assigning themes and aggregating similar clusters. This process gave way to a number of \textit{reasoning patterns} for privacy rating choice, which we enumerate in the following with representative feedback quotes:

\begin{itemize}
    \item \textbf{Presence of direct identifiers and named entities}: \say{\textit{I was generally looking around for names, locations, companies, usernames, hashtags, etc., that could possibly give the text a way to search for the person in question.}}
    \item \textbf{General possibility of being identified}: \say{\textit{I was looking at the questions from a perspective of if the question had any information where others could find out who the person is posting the question or could figure our where they live or where they work for example. Seeing how people can 'dox' themself without knowing sort of thing.}}
    \item \textbf{Privacy based on topic / domain sensitivity}: \say{\textit{Mostly the topics. Mental health seemed more private to me, and things like reviews or public posts less sensitive.}}
    \item \textbf{Very personal vs. identifiable}: \say{\textit{If you cannot identify the person, I always chose harmless, no matter what kind of information the text contained.}}
    \item \textbf{Private or public?}: \say{\textit{I felt as some things are private and are meant to be private.}}
    \item \textbf{Risk of harm}: \say{\textit{I judged by the degree of embarrassment the [user] would likely feel if their message wasn't anonymized.}}
\end{itemize}

Thus, we see that many of the reasons provided by the survey participants revolve around either direct identifiers, or also topics that \say{seem} sensitive, risky, or potentially harmful. We observed many comments of \say{putting oneself in another's shoes}, especially in considering the harm of revealing oneself via indirect identifiers.

The LLM responses are thorough but display much lower variety, where nearly all reasons given by any LLM address two major aspects: \textit{sensitivity} and \textit{identifiability}. Often, an LLM would reason by explicitly extracting indirect identifiers (e.g., health history, traumatic story, personal feelings, etc.), as well as direct identifiers, if present. Based on the level of sensitivity of these, combined with the perceived risk of identifiability, the LLMs would provide their score accordingly. A representative LLM reason is given below:

\begin{quote}
    \small
    \textit{The user\_text contains specific details about a meeting, including dates, locations, and names of individuals involved. While it does not include highly sensitive personal information, it does contain identifiable information that could be linked to specific individuals and organizational activities. Therefore, it is somewhat private but not extremely so.}
\end{quote}

In the juxtaposition of human and LLM reasoning patterns, we find that while humans may express greater creativity in the evaluation of privacy, LLMs are more objective and consistent, following directly the instructions provided in the prompt (and the guidelines of the privacy scale). While this consistency may not always be desirable in representing the diverse ways in which a human might approach the task, it does lead to a reliable source of reasoning, and this is reflected in the inter-LLM agreement scores. Above all, this distinction provides evidence for the importance of the privacy definition, i.e., the construction of the prompt, which clearly influences and confines the way in which LLMs will complete the task. Taking the two overarching perspectives into consideration, we begin to observe a general formula for an understanding of text privacy, which lies at the intersection of \textit{sensitivity} and \textit{identifiability}. 

\subsection{How do we evaluate privacy in text?}
We contextualize our results in the scope of privacy-preserving NLP and working with sensitive text data, and we analyze what our findings teach us about how privacy evaluations can be conducted to take a human-centered, yet efficient and cost-effective approach.

A clear lesson comes with the evidence that privacy evaluation largely depends on the domain from which user-written texts originate. Text datasets containing personal messages about health conditions, emotional feelings, or personal life situations are clearly more important in capturing human perceptions of privacy than, for example, movie reviews or Twitter posts. Thus, in the evaluation of privacy in textual data, the general domain is an important first consideration in debating the need for human evaluation.

Beyond this, future evaluations of privacy in NLP, specifically on the sensitivity of the data itself, must consider the trade-offs between full human evaluation and leveraging LLM-as-a-Judge as a privacy evaluator. In addition to the benefits and limitations discussed above, we highlight the remaining major factor of cost. For the capturing of 13,540 human privacy ratings (677 participants $\times$ 20 texts), we incurred a cost of \textsterling 2031, whereas for the 16,250 LLM responses (13 LLM $\times$ 250 texts $\times$ 5 runs), the costs were significantly less, estimated to be less than \$20 including negligible hardware costs for the open-source models. This difference is significant, and given the generally high LLM-human agreement scores, a strong case is made for LLM-as-a-Judge from a resource perspective.

There are other factors to consider, however, in conducting privacy evaluations on textual data. An immediate concern may arise in cases where non-public datasets are used, in which achieving strong evaluation results may necessitate sending this data to third-party model providers, which may not be desirable. As another point, running large open-source models locally may be prohibitive, in contrast to Prolific (in our case), which presents no hardware costs. Finally, the major limitation of conducting LLM-as-a-Judge evaluations lies in the inability to capture or accurately represent diverse human opinions, especially in the consideration of privacy, which is incredibly personal and human-centered. Thus, solely relying on LLMs for evaluation may prohibit deeper investigative studies into personal privacy opinions, or rather, studying what inherently builds our inner theories of privacy (in text or beyond).

\section{Conclusion}
We investigate the use of LLMs in the evaluation of privacy in textual data, and we empirically measure the alignment of privacy perceptions between LLMs and human annotators. With a dataset of 250 texts representatively sampled from 10 datasets of user texts, we capture the privacy ratings of 13 LLMs and 677 human participants, providing the basis for the calculation of inter-LLM, inter-human, and human-LLM agreement levels across all rated texts. Our findings point to the general usability of LLMs as an accurate global approximation for human privacy sentiment, while also highlighting the limitations in using generalized, non-representative LLMs for an inherently personal, human-centered task.

\textit{Limitations.}
We acknowledge that our work is limited by the selected datasets, in that they may not be fully representative of the universe of user-written texts. Similarly, our selected LLMs only represent a fraction of the currently available and widely used models, from which we chose the most popular proprietary and open-source models. Additionally, we do not account for the possibility of data contamination, where some of the used LLMs may have already seen these text examples in other contexts (as they are public datasets). We also do not experiment with one- or few-shot prompting, which could reasonably improve the alignment of LLM and human privacy perceptions. Our utilized prompts are also limited by the operationalized five-point privacy scale, which still necessitates interpretation, both in the case of humans and LLMs.

Our survey is limited in its geographic scope, where a considerable majority of the participants come from Europe or Africa. Finally, we are limited by our provided definition of privacy (sensitivity) and the translation of this definition into the prompts and survey instructions. We caution that our results should only be interpreted under this definition, and the generalizability of our findings to other definitions of privacy remains as future work.

\textit{Future Work.}
As follow-up work, we recommend in-depth studies that extend beyond agreement metrics to investigate \textit{what exactly} (linguistically and semantically) in textual data humans perceive to be private. This includes studies of cross-cultural perceptions of privacy, and how these diverse perspectives can be comprehensively represented in the evaluation of privacy-preserving NLP. Additionally, future research could also focus on the construction of lightweight, yet accurate models for evaluating privacy sensitivity in text, of course, with a basis in human-annotated data.


\newpage
\bibliographystyle{ACM-Reference-Format}
\bibliography{sample-base}

\appendix

\onecolumn
\section{Example Texts}

\begin{table}[hbtp]
\small
\caption{Examples of user texts from each of our 10 selected datasets, along with the average LLM and human privacy rating.}
\resizebox{\linewidth}{!}{
\begin{tabular}{l|p{0.9\linewidth}|c|c} 
\textbf{Dataset} & \textbf{Text} & \textbf{LLM Avg} & \textbf{Human Avg} \\ \hline
BAC & Sorry but please tell me that you (abby) were being sarcastic about Becca. I really really really want to like her but it isn't working. & 2.7 & 2.8 \\
EE & Please move all the following counterparties(financial \& natural gas trades only)deals from ENA-FT-WT-SOCAL book to the BANKRUPTCY book: HESSENESER BRIDGELIGASMAR CONSUMERS CORNERSTPROL P IGIRES OCCIDENTENEMAR PRAXAIR PSEGENERES SEMPRAENETRA Please move all the following counterparties (financial \& natural gas trades only) from NG-PRICE book to the BANKRUPTCY book: CARGILL CONTOURENECO DEUTSCHEBANAKT DUKEENENGLLP NATIONALENETRA N-IL-GAS PRIMARYNATRES WESTPORTOILGAS WPSENGRYSVC We are heading out shortly, so if you have any questions, please call me at home 281-486-7026 or cell 713-819-0765, or Darron Giron at home 281-304-8303 or cell 832-524-6091. Thanks for your help. & 4.2 & 3.8 \\
MHB & Lately I have wished there was a magic switch to turn off my anxiety. It's just so hard to deal with. Does anyone else feel like this? & 2.8 & 2.3 \\
RMHP & since i realized it's ocd, i feel like i have no control over my own brain, it keeps sending me very bad intrusive thoughts, i used to watch criminal psychology and true crime videos on youtube, but there was one that i watched that i keep getting intrusive thoughts about, cause it was so brutal. i read fanfiction and stuff, and im reading a batman one recently, where the joker was killed and cut up and i remember that story cause of it. im afraid of dating in general cause im afraid im gonna hurt someone. i also get intrusive thoughts about animals, and i think i look at the private parts cause of it, i have a weird thing with looking at private parts its weird and i hate it, not just with animals with human beings too, i used to want a dog, but now i dont know cause im afraid of these thoughts. i get sexual intrusive thoughts about my family and that also hurts me, cause i feel super numb to all these thoughts now, i think im just equally depressed. i dont think im in control anymore and i hate it. i hate thinking im gonna harm anyone, i hate knowing that when im awake in the middle of the night and go to the kitchen the thoughts of stabbing my mom in her sleep pops up. or then stabbing myself instead, or carving my own eye balls out. i need this to stop cause i want control over my life again, i don't want to feel depressed and hopeless anymore. i feel like a monster half of the time, i dont feel like myself at all. i dont remember know what it feels like to be okay. i was okay before the pandemic but being stuck in a constant loop of not doing anything kills me, i applied for a job a couple of weeks ago, didn't hear back yet, i might never hear back. i need to start the ged program but i feel unmotivated. i go to therapy on the 21st for my first actual appointment, i did the orientation and the introductory already. but i also don't know how to talk about this outloud and come to terms with the thoughts. & 4.7 & 4.1 \\
TR & Brilliant Fitters - Average Company: After a sometimes surreal planning stage the end result was worth it. Betta Living's fitters are amazing. Two days of chaos (not their fault) and I have an amazing new Bedroom. [Name] \& [Name] were punctual, pleasant, tidy and took great care with the installation and my property; they were a pleasure to have around. The 5 Stars are for them Betta Living need to review their internal communication between departments. I found sales, surveyor and finance were not singing from the same Hymn sheet......room for improvement here. 3 Stars are appropriate. NAME & 2.9 & 2.0 \\
TW & Boundless gratitude for all our Veterans today. And every day. Thank you for your service. And thank you to all your families as well as everyone plays their part. Love you guys. Thank you. Forever proud \#veteransday & 1.5 & 1.4 \\
YR & Guapos Tacos are served street-style, with the protein taking center stage and chopped onions and cilantro plus light sauces adding to the flavor. My first experience was at Amis industry night last September. There I tried tacos de lengua (tongue), tacos de carnitas and veggie tacos. The standouts were the lengua and veggie. The tongue was tender like a filet and the accompanying sauce had a tangy flavor. The veggie taco had flavorful, marinated grilled veggies. My second taste of Guapos Tacos was at the Vendy Awards. The grilled mahi mahi taco with a generous splash of hot sauce (from the selection of sauces on the ledge built into the the truck) was so delicious, I gladly waited in line again for seconds later in the day. The truck is colorful and fun and the guys who work inside are personable and handle crowds well. Two thumbs up for this tasty taco truck! & 1.5 & 1.5 \\
RC & I am not proud one bit at all. I cannot wait until I have the opportunity to leave this miserable country. You see, my dad is my absolute best friend. America sent my dad over to Vietnam when he was 19. He was sprayed with a pesticide that caused Agent Orange Toxicity. This in turn caused Parkinson’s disease and severe dementia to the point that I had to put him in a home. I am lonely, sad, and I miss eating dinner with my best friend every night. If the US wouldn’t have gone over to Vietnam, my dad would be in the conditions he is in. The US had no business over there and should have never gone. Too many people were hurt and killed as a result. I am not proud to be from the US. I am ashamed. I am ashamed of our politics and the way we act. I’m ashamed of how others see us. I’m ashamed at how greedy we all are. I am not a proud American. When my dad passes, I plan to migrate to another country. I would now if it was possible. I hate it here. I. Am. Not. Proud. & 3.8 & 3.0 \\
RLA & To start off with, I work for the Florida Department of Corrections. I have been with them for about seven years. In a timeframe of approximately Six months I have been assaulted with bodily fluids three times by the same inmate. The first incident involved the inmate taking control of the cell door flap and throwing feces on me. In times where an inmate has assaulted an officer, they will be separated in the form of either moving the inmate to a different location or reassigning me to another building. We were not separated and about four months later while the inmate was in a recreation cage, I was conducting count when he spit in my face. At this time I was taken to the emergency room and placed on medications for potential HIV exposure for fourteen days. We were separated after this incident as I was moved to another assignment. Now two months later upon getting my assignment I was told to report to the same building that the inmate in question was housed in. Within twenty minutes of being there the inmate defeated the security device that locks his cell door flap and threw feces into my eyes. Again I was sent to the emergency room and placed on medication for potential HIV exposure. I feel as though the second and third time I was assaulted could have been easily avoided and it wasn't. I don't know if I have a case or anything so I'm coming here to get any advice or information. These incidents have affected me mentally, physically in the form of this medication, and it is affecting my family. Any help is greatly appreciated. & 4.3 & 4.1 \\
MQ & My grandmmother is 81 yrs old. Checked her bp several times over a 3hr period and always got 190/90. Doctor gave her CAPTOPRIL 10 minutes ago. Her hands and feet are cold. Shes weak and cant stand up. She can move her arms and feet and didnt have a hard time in the bathroom 3 hrs ago. Pls. advise. Tks. & 3.9 & 3.3
\end{tabular}
}
\label{tab:examples}
\end{table}

\end{document}